\documentclass{article}
\usepackage{ijcai20}

\usepackage{times}
\usepackage{soul}
\usepackage{url}
\usepackage[hidelinks]{hyperref}
\usepackage[utf8]{inputenc}
\usepackage[small]{caption}
\usepackage{graphicx}
\usepackage{amsmath}
\usepackage{amsthm}
\usepackage{amssymb}
\usepackage{booktabs}
\usepackage{algorithm}
\usepackage{algorithmic}
\usepackage{subfigure}
\graphicspath{ {fig/} }

\title{Following Instructions by Imagining and Reaching Visual Goals}

\author{
John Kanu$^1$
\and
Eadom Dessalene$^1$\and
Xiaomin Lin$^2$\and
Cornelia Ferm\"uller$^3$\And
Yiannis Aloimonos$^1$
\affiliations
Department of Computer Science, University of Maryland, College Park
\emails
$^1$\{jdkanu,edessale,yiannis\}@cs.umd.edu,
$^2$xlin01@umd.edu,
$^3$fer@umiacs.umd.edu
}
\begin{document}

\maketitle

\begin{abstract}
%Visual images implicitly encode scene structure relevant to the completion of tasks, such as the shape, color, and location of objects to be interacted with. When performing tasks, humans perform spatial reasoning on the basis of this structure in order to generate sub-goals, and fulfill sub-goals using general-purpose skills.
While traditional methods for instruction-following typically assume prior linguistic and perceptual knowledge, many recent works in reinforcement learning (RL) have proposed learning policies end-to-end, typically by training neural networks to map joint representations of observations and instructions directly to actions. In this work, we present a novel framework for learning to perform temporally extended tasks using spatial reasoning in the RL framework, by sequentially imagining visual goals and choosing appropriate actions to fulfill imagined goals. Our framework operates on raw pixel images, assumes no prior linguistic or perceptual knowledge, and learns via intrinsic motivation and a single extrinsic reward signal measuring task completion. We validate our method in two environments with a robot arm in a simulated interactive 3D environment. Our method outperforms two flat architectures with raw-pixel and ground-truth states, and a hierarchical architecture with ground-truth states on object arrangement tasks.
\end{abstract}

%\footnote{Code is available at \href{https://github.com/jdkanu/IFIG}{\tt{https://github.com/jdkanu/IFIG}}.}

\section{Introduction}

Building agents that can follow instructions in physical environments is a longstanding challenge in the development of artificial intelligence, originally introduced as SHRDLU in the early 1970s \cite{winograd1971procedures}. While traditional methods for instruction-following typically assume prior linguistic, perceptual, and procedural knowledge in order to execute instructions \cite{paul2017grounding,guadarrama2013grounding,misra2016tell,tellex2011understanding}, recent works have shown that deep reinforcement learning (RL) may be a promising framework for instruction-following tasks in navigation \cite{hermann2017grounded,chao2011towards} and non-robotic manipulation \cite{misra2017mapping,bahdanau2018learning,jiang2019language}. Naturally, instructions often represent temporally extended tasks, whose successful execution requires language grounding, structured reasoning, and motor skills. In order to design agents that exhibit intelligence and learn through interaction, it will be useful to incorporate cognitive capabilities like imagination, one of the hallmarks of human-level intelligence \cite{tenenbaum2018building}. Learning to effectively ground language instructions and the skills to achieve them in the deep RL framework is still an open challenge. In this work, we address the need to incorporate explicit spatial reasoning to accomplish temporally extended tasks with minimal manual engineering.

%TODO clarify this sentence: "while traditional methods for instruction-following ... recent works have shown that ..." why does deep RL stand in contrast

\begin{figure}[!t]
    \centering
    \includegraphics[width=0.39\textwidth]{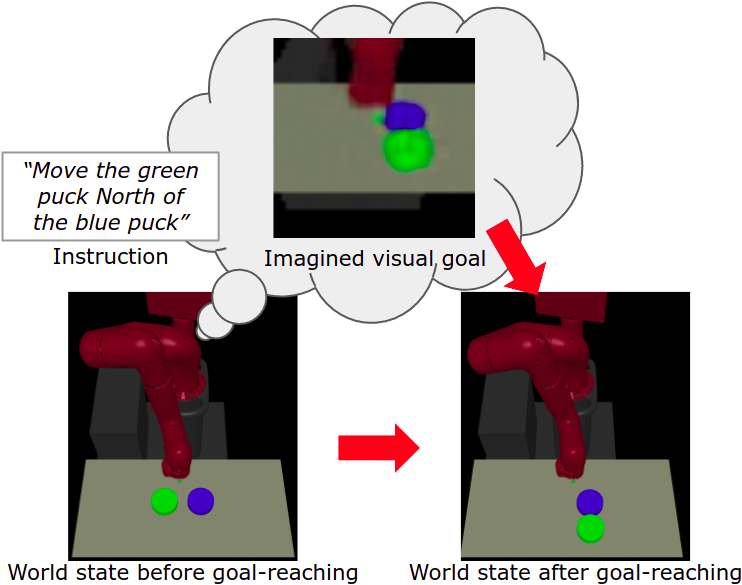}
    \caption{Our agent learns to execute instructions by synthesizing and reaching visual goals from a learned latent variable model.}
    \label{fig:fig_intro}
\end{figure}

%TODO switch NORTH and SOUTH in all instructions and figures (they're just labels) - makes the figures more natural and understandable

%TODO add arrow from instruction to visual goal in Figure 1

%Moreover, studies in developmental cognitive science suggest that the acquisition of various words is closely linked to the learning of their underlying concepts in the physical world \cite{gopnik_meltzoff_1984}, suggesting.

In several previous works, models are trained to ground language using a multimodal state representation \cite{janner2018representation,chaplot2018gated}, and acquire skills through a flat policy that maps this state to an action. However, such techniques are based on inductive biases that limit modeling flexibility. \cite{jiang2019language} employ hierarchical reinforcement learning (HRL) to train a hierarchy of policies to perform temporally extended tasks through goal-directed behavior, with a high-level policy issuing goals to a low-level policy in the form of language. While language is an expressive modality, a low-level policy conditioned on language resembles a flat instruction-following policy with limited language grounding mechanisms. One possibility to ground language into the context in the HRL framework is to use hard-coded abstractions \cite{peng2017deeploco,heess2016learning,konidaris2007building}, though this technique limits the flexibility to perform arbitrary tasks specified by language. One can instead use raw sensory signals such as images, representing transformations of the entire state; however, photometric loss functions such as Euclidean distance often do not correspond to meaningful differences between states \cite{ponomarenko2015image,zhang2018unreasonable} and are thus ineffective for training low-level policies. Another possibility is to learn a structured latent representation of the image. As demonstrated in \cite{nair2018visual,pong2019skew}, motor policies can be trained on image embeddings with reward based on distances in latent space.

Building upon the HRL framework and methods for goal-conditioned, visual RL \cite{nair2018visual,pong2019skew}, we introduce a framework consisting of three main parts: (1) a state representation module, which learns a structured latent representation of image states, (2) a goal-generating module, which learns instruction-conditional structured state transformations to sequentially imagine goal states while executing the instruction, and (3) a goal-reaching module, which learns the skills to move between arbitrary states. Learning structured latent representations according to an image reconstruction objective allows the agent to learn to represent the information in the state completely self-supervised, while learning to generate and reach goals using separate objectives allows us to decouple the visual grounding of the instruction from the lower-level acquisition of skills. Imagining transformations to the visual state also achieves a form of task-directed spatial reasoning. We call this method Instruction-Following with Imagined Goals (IFIG).

%in which an agent learns visual state representations for goal-reaching, and learns to sequentially imagine and reach visual states corresponding to the execution of the instruction given the current visual context. The architecture

We conduct experiments on physical object-arrangement tasks inside a simulated interactive 3D environment, converting instructions to continuous control signals on a robot arm. We benchmark our model against a flat policy with raw-pixel states, a flat policy with ground-truth states, and a hierarchy of policies with ground-truth states. Experiments show that our model outperforms all three, with accuracy measured by distance between each object and its goal position. Through visualization of goal reconstructions, we show that our model has learned instruction-conditional state transformations that suggest an ability to perform task-directed spatial reasoning.

\section{Related Work}

\subsubsection{Language grounding in robotics}

Traditional methods exploit the compositional structure of language to generate action plans. Several methods infer action plans using pre-defined templates for semantic parsing, object localization, and modeling of spatial relations \cite{paul2017grounding,misra2016tell,guadarrama2013grounding,tellex2011understanding}. These methods require hand-crafted representations of the world model, object relations, and object attributes such as geometry and color, which can limit the agent's ability to specify arbitrary goal states that are not captured by the model. To our knowledge, our work is the first to provide an approach to visually ground language for object manipulation using RL, without assuming prior linguistic or perceptual knowledge.

\begin{figure}[h]
    \centering
    \includegraphics[width=0.42\textwidth]{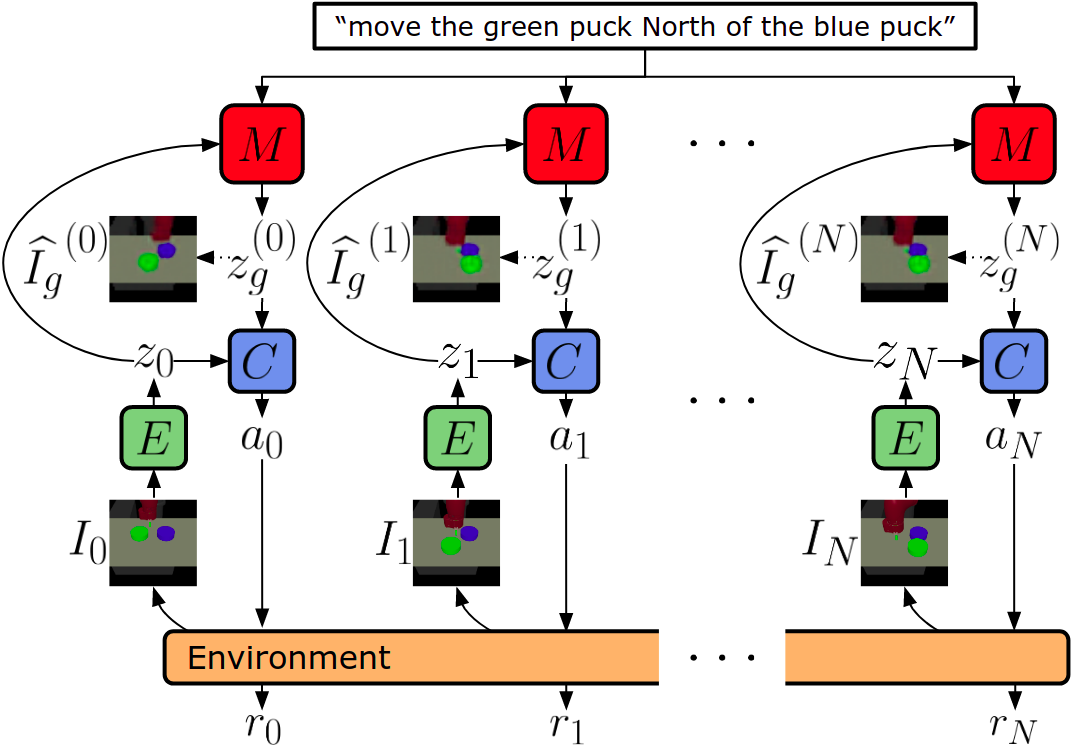}
    \caption{At each time $t$, we embed the image $I_t$ of the environment into a latent state $z_t$. Given $z_t$ and an instruction, the meta-controller generates a latent goal state $z_g^{(t)}$ (which can reconstruct an image $\widehat{I_g}^{(t)}$), and the controller generates an action $a_t$, yielding reward $r_t$. This process repeats until the episode terminates.}
    \label{fig:fig_main}
\end{figure}

\begin{figure*}[t!]
    \centering
    \includegraphics[width=0.82\textwidth]{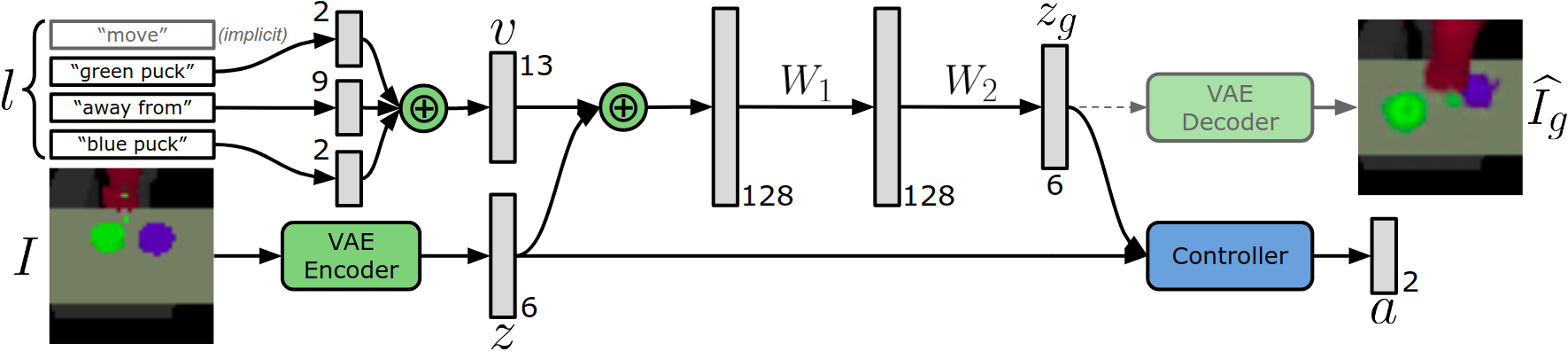}
    \caption{Our architecture for generating motion commands given instructions and an image of the environment. For simplicity, instructions are concatenations of three word embeddings. The architecture generates and reaches a visual goal representing the next step of the task.}
    \label{fig:fig_mc}
\end{figure*}

\subsubsection{Mapping instructions to actions using deep RL}

In recent years, deep RL has been applied to instruction-following tasks for navigation and non-robotic manipulation in simulated 2D and 3D environments. Agent architectures typically consist of a mixing module which provides a multimodal representation of the instruction and state, and a policy network which learns to map the combined state representation to an action. Simple concatenation of instruction and state is used in \cite{mei2016listen,misra2017mapping,hermann2017grounded}. More sophisticated methods for visual grounding of language have also been explored. \cite{chaplot2018gated} use an attention mechanism based on multiplicative interactions between instruction and image state. \cite{janner2018representation} learn to generate a convolutional kernel from the instruction, which is used to convolve the image before passing it to a policy. \cite{jiang2019language,hu2019hierarchical} train a high-level policy to issue language instructions to a low-level policy, which learns to generate actions from language with limited language grounding mechanisms. \cite{prabhudesai2019embodied} train object detectors and generative networks to map instructions and an image to a 3-dimensional visual feature representation of the desired state, but their method requires language supervision from a semantic parser, and their framework only supports spatial translations of objects. In contrast to these works, we train a generative network to provide a latent state space from raw pixels, to allow a high-level policy to produce arbitrary transformations in the latent space conditioned on language instructions, and a low-level policy to acquire general-purpose skills in achieving arbitrary state transformations. We make no assumptions on prior linguistic and perceptual knowledge.

\section{Preliminaries}

Our method combines methods for unsupervised representation learning, goal-conditioned reinforcement learning \cite{nair2018visual,pong2019skew}, and text-conditional image transformations. We briefly review the methods we employ.

\subsection{Goal-conditioned reinforcement learning}

Our meta-controller and controller are each trained using goal-conditioned reinforcement learning, where goals are instructions for the high-level policy and states for the low-level policy. The goal-conditioned RL problem considers an Augmented Markov Decision Process defined by the tuple $(S,G,A,T,R,\gamma)$, where $S$ is the state space, $G$ is the goal space, $A$ is the action space, $T:S\times A\times S\rightarrow [0,\infty)$ represents the probability density of transitioning from $s_t\in S$ to $s_{t+1}\in S$ under the action $a\in A$, $R:S\times A\times G\rightarrow [r_{\textrm{min}},r_{\textrm{max}}]$ represents the scalar reward for each transition under a given goal, and $\gamma\in [0,1)$ is the discount factor. The objective is to learn a policy $\pi(a_t|s_t,g)$ that maximizes the expected discounted return $R_t = \mathbb{E}[\sum_{i=t}^{T}\gamma^{(i-t)}r_i]$, where $r_i=R(s_i,a_i,g)$.

\subsection{Reinforcement learning with imagined goals}
\label{rig}

Reinforcement learning with imagined goals (RIG) is a framework for jointly learning a latent representation of raw sensory states and a policy for reaching arbitrary latent goal states \cite{nair2018visual}. We summarize the algorithm as follows. First, an exploration policy is used to collect a dataset $D$ of visual observations. In our case, we use Skew-Fit \cite{pong2019skew} for exploration, a self-supervised state-covering method which provides a formal exploration objective that maximizes the entropy of the goal distribution. Second, a beta variational autoencoder ($\beta$-VAE), which has been shown to learn structured representations of high-dimensional data \cite{kingma2014stochastic}, is trained on $D$ to learn an embedding of raw pixel images. The VAE consists of an encoder $q_\phi(z|s)$ that models the probability of latent distribution $z$ corresponding to the visual state $s$ given parameters $\phi$, and a decoder $p_\psi(s|z)$ that models the probability of visual state $s$ corresponding to the latent distribution $z$ given parameters $\psi$. The encoder $q_\phi$ and decoder $p_\psi$ are approximated by neural networks mapping between image states and latent states. After training the VAE, a policy $\pi(z,z_g)=a$ is trained to generate actions $a$ that move an agent from $z$ to $z_g$, where $z_g$ is a latent goal sampled from the latent variable model. Reward is computed as $R(s,s_g)=-||e(s)-e(s_g)||_2=-||z-z_g||_2$, the Euclidean distance between the current latent state and the goal latent state.

\section{Problem Formulation}

We address the problem of instruction-following in interactive environments, where an autonomous agent is issued a task in the form of an instruction, and must produce actions to complete the task. Assume the agent exists inside an episodic environment. At the beginning of each episode, the environment is randomly initialized, and an instruction $l$ is generated. At each time step $t$, the agent is given the instruction $l$ and a raw pixel image $I_t$ of the environment, and generates an action $a_t$. We consider the case that the instruction is embedded as a fixed-length vector $v$ by concatenation. We assume the existence of a function $R_l : S\rightarrow [r_{\textrm{min}},r_{\textrm{max}}]$ which outputs the agreement between the instruction $l$ and state $s_t$. The episode terminates when either the agreement is maximized or the maximum number of steps is reached. The objective is to learn a policy $\pi(a|s,l)$ that maximizes the expected return, which corresponds to a successful completion of the task specified by the instruction within the time limit.

\section{Method}

In this section, we present our framework for training an agent to perform continuous control conditioned on language instructions and raw pixel images, by integrating a structured state representation module, a high-level policy that generates visual goals from instructions and observations, and a task-agnostic low-level policy that performs control to reach visual goals. We refer to this framework as \textit{Instruction-Following with Imagined Goals} (IFIG, Figures \ref{fig:fig_main} and \ref{fig:fig_mc}).

%We factorize the agent into a \textit{meta-controller}, which generates a goal given the instruction and current state, and a \textit{controller}, which maps the current state and goal to an action. The meta-controller is trained to map an image and instruction to a goal image in latent space. During test-time, the meta-controller will generate a goal given the current state and instruction, and the controller will map the current state and goal state to a sequence of actions. 

In Section \ref{statereplearn}, we summarize a method for learning state representations $z$ from visual observations $s$ fully unsupervised \cite{nair2018visual}. In Section \ref{goalreachpol}, we summarize a method for training a low-level, task-agnostic goal-reaching policy $\pi_l (a|z_t,z_g)$, also fully unsupervised \cite{pong2019skew}. In Section \ref{goalgenpol}, we discuss our method for training a high-level policy $\pi_h (z_g|z_t,v)$ to generate task-directed structured transformations of visual inputs given instruction embedding $v$ and context $z_t$, using the low-level policy, which we assume is already trained. For convenience, we use the terms \textit{low-level policy}, \textit{goal-reaching policy}, and \textit{controller} interchangeably, and the terms  \textit{high-level policy}, \textit{goal-generating policy}, and \textit{meta-controller} interchangeably.

\subsection{State representation learning}
\label{statereplearn}

In order to map instructions and states to goals, and current and goal states to actions, we benefit from finding a suitable state representation. The state representation must reflect the factors of variation in the scene relevant to instruction-following tasks, such as the number, position, and attributes of objects such as shape and color. Raw pixel images solve this problem, but using pixel images as goals can be challenging, since photometric loss functions like Euclidean distance usually do not correspond to meaningful differences between states \cite{ponomarenko2015image,zhang2018unreasonable} and are thus ineffective for computing an intrinsic reward signal. Variational autoencoders (VAEs), however, have been shown to learn structured representations of high-dimensional data \cite{kingma2014stochastic} and have been used to represent states for robotic object manipulation \cite{nair2018visual}.

We train a $\beta$-VAE by self-supervised learning according to the RIG algorithm \cite{nair2018visual}, and use the latent space to represent world states. This allows us to compute reward for goal-reaching using distances in latent space when training the low-level policy, and provides a latent distribution for the goal space when training the high-level policy. The high-level policy can learn to specify instruction-conditional transformations of the environment structure by generating goals in the continuous latent space and improving the policy parameters over each episode using an RL method suitable for continuous action spaces.

Following the approach of Nair et al. (2018), we collect a dataset of state observations $\{s^{(i)}\}$ from the execution of a random policy \cite{pong2019skew}, and train a $\beta$-VAE on the dataset. We subsequently fine-tune the $\beta$-VAE. We embed the state $s_t$ and goals $s_g$ into a latent space $\mathcal{Z}$ using an encoder $e$ to obtain a latent state $z_t=e(s_t)$ and latent goal $z_g=e(s_g)$. We use the mean of the encoder as the state encoding, i.e. $z_t=e(s_t)\triangleq \mu_\phi (s_t)$. See Section \ref{rig} for more details.

\begin{figure}[t]
    \centering
    \includegraphics[width=0.47\textwidth]{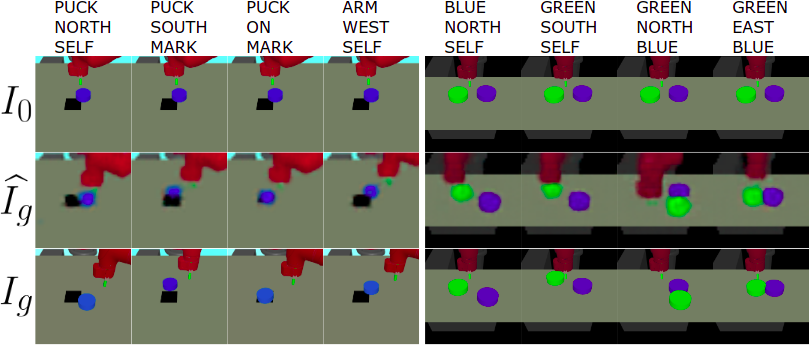}
    \caption{Illustrations of selected instructions from the single object and multiple object arrangement tasks. Top row is the initial image $I_0$. Middle row is the reconstruction $\widehat{I_g}$ of the model's final latent goal state $z_g$. Bottom row is the image of the correct final state $I_g$.}
    \label{fig:fig_instrs_all}
\end{figure}

%TODO update PUCK SOUTH MARK with correct goal I_g

\subsection{Goal-reaching policy}
\label{goalreachpol}

To reach goals, we train a low-level policy $\pi_l(a|z,z_g)$ that generates actions to move from a given current state $z$ toward a goal state $z_g$, according to the RIG framework \cite{nair2018visual}. This process is self-supervised. The reward function is given by the function $R(s,s_g)=-||e(s)-e(s_g)||_2=-||z-z_g||_2$, the Euclidean distance between the current latent and the latent goal. See Section \ref{rig} for more details. With the trained low-level policy $\pi_l$, the agent can attempt to reach arbitrary goals in the latent distribution $Z$.

\begin{figure*}[htbp]
\centering
\includegraphics[width=1.0\textwidth]{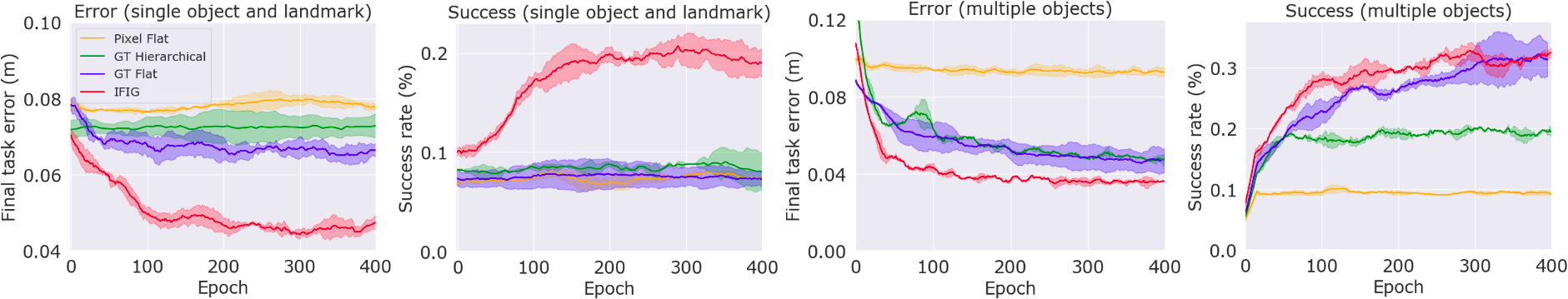}
\caption{Cumulative distance to goals for all instructions. To normalize training over different temporal scales, we divide flat durations by 10, since 1 step of the meta-controller corresponds to 10 steps of the controller. Plots show mean and standard deviation across 3 seeds.
}
\label{fig:fig_results}
\end{figure*}

\subsection{Goal-generating policy}
\label{goalgenpol}

To generate goals from instructions, the goal-generating policy $\pi_h$ maps the instruction and the current state to a goal state, which is sent to the goal-reaching policy, as seen in Figure \ref{fig:fig_mc}. We approximate the optimal policy $\pi_h$ with a fully-connected multi-layer perceptron (MLP) with two layers of weights $W_1$ and $W_2$. The process of sequentially generating goals is described as follows. At each time step, the raw-pixel image $I_t$ is embedded into a latent vector $z_t$ by the encoder network. We assume the existence of a function $L : l\rightarrow \mathbb{R}^N$, which embeds the instruction $l$ to a fixed-length vector $v$ of dimensionality $N$. Thus, the joint state representation of $I_t$ and $l$ is simply the concatenation of the embeddings $z_t$ and $v$, i.e. $s_t = [z_t;v]$. Finally, $\pi_h$ maps $s_t$ to a task-directed goal $z_g$ in latent space, which is sent to the goal-reaching policy $\pi_l$. Once $\pi_l$ reaches the goal, the process is repeated, unless the instruction is completed or the episode terminates. See Figures \ref{fig:fig_main} and \ref{fig:fig_mc} for an illustration.

We train the meta-controller using soft actor-critic (SAC), a popular algorithm for learning policies with continuous action space that has demonstrated good performance on similar tasks \cite{haarnoja2018soft}. A continuous scalar reward signal $r$ is issued by the environment based on the Euclidean distances between the actual and desired object positions:

\begin{equation}
    R(s,l)=\sum_{o\in O}-||x_o - g_o||_2
    \label{reward}
\end{equation}

where $O$ is the set of objects, $x_o$ is the current position of $o$, and $g_o$ is the goal position of $o$ under instruction $l$. The negation yields higher reward to states that are closer to goals.

%It is important to note that, since reward is maximized only if the desired goal state is reached, the ability of the goal-generating policy to produce a goal state representing the successful completion of the instruction is dependent on the ability of the controller to reach the desired goal state.

One common issue with hierarchical policies comes from updates to the low-level policy during training, which necessitates off-policy corrections \cite{nachum2018data}. Since we split the training into two disjoint phases, our method does not require off-policy corrections.

\section{Experiments}

We are interested in training agents to perform task-directed spatial reasoning with minimal manual engineering, by training a policy to transform structured visual observations according to symbolic instructions using RL. We aim to answer the following two questions:

\begin{enumerate}
    \item How effective is our approach in comparison with methods that do not make any explicit objective for spatial reasoning or goal-setting?
    \item How effective is the visual modality for representing sub-goals for instruction-following, in comparison with other modalities?
\end{enumerate}

We devise a set of experiments to address these questions. For the first question, we compare an IFIG agent to two flat agents, one of which operates on raw images and another that operates on ground truth states, meaning its observations are the ground truth positions of objects in the environment. For the second question, we compare an IFIG agent to a hierarchical agent with ground truth states. Section \ref{environment} discusses the 3D interactive environment, the object arrangement tasks, and instruction representation. Sections \ref{baselines} and \ref{hyperparams} discuss each approach in detail.

\subsection{Environment}
\label{environment}

A natural test environment occurs when tasking a robotic arm to physically manipulate objects, where an instruction specifies a desired object arrangement. We consider the problem of visual object-manipulation, where an agent controls a robotic arm inside an interactive, episodic 3D environment. At the beginning of each episode, the objects and arm configuration are randomly initialized, and an instruction $l$ is generated, describing a set of spatial relations among objects in the environment to be achieved by the agent. At each time step, the robotic arm produces an action specifying the velocity of its end-effector. The environment updates its state according to $a$, by updating the joint configuration of the robot arm via inverse kinematics, performing collision detection and response, and updating the position of all physical entities.

We implement environments for two tasks inside the MuJoCo simulator with a Sawyer arm and continuous control: single object arrangement and multiple object arrangement. Movable pucks, 0.02 m in height and 0.08 m in diameter, are placed on a flat table in front of the robot. A landmark is indicated by a square on the surface of the table. The agent is given an RGB pixel image of the task space from a static viewpoint. The action space is 2-dimensional in the $xy$-plane, describing the change in the end-effector position at each time step. At the beginning of each episode both puck positions are sampled from uniformly random distributions of positions within a rectangular region on the table. The environment is reset after the instruction is completed or after 250 steps. See Figure \ref{fig:fig_instrs_all} for a visualization of the environment and both tasks.

\subsubsection{Single object arrangement with landmark}

In this task environment, the agent is tasked with arranging a single puck on the table, with a landmark on its surface. Both the puck and the landmark are randomly initialized. We issue 15 arrangement instructions, each describing a change in the puck position relative to its current position (e.g., \textit{``move blue puck North''}), a change in the puck position relative to the landmark (e.g., \textit{``move blue puck onto landmark''}), or a change in the arm position (e.g., \textit{``move arm East''}).

\subsubsection{Multiple object arrangement}

In this task environment, the agent must arrange two pucks. Both pucks are randomly initialized on their respective sides. We issue 22 arrangement instructions, each describing a change in the position of either puck relative to its current position (e.g., \textit{``move blue puck West''}), a change in puck position relative to the other puck (e.g., \textit{``move blue puck away from green puck''}), or a change in arm position (e.g., \textit{``move arm East''}). For simplicity, we do not issue instructions to move the puck to the opposite side of the other puck.

\subsubsection{Instructions}

We create a vocabulary of 12 tokens referring to the blue puck, green puck, black landmark, arm, cardinal directions (North, South, East, and West), binary relations (representing the prepositions \textit{``on''}, \textit{``near''}, and \textit{``away from''}), and a self-identifier. Each instruction is uniquely tokenized by a fixed-length sequence of tokens $l=(l_1,l_2,l_3)$, where $l_1$ identifies the object that should be moved (blue puck, green puck, or arm), $l_2$ identifies the spatial relation, and $l_3$ identifies the reference object (e.g., \textit{``move green puck North of blue puck''} becomes \texttt{[GREEN,NORTH,BLUE]}). $l_3$ takes the value of the self-identifier token for instructions like \textit{``move blue puck North''}.

We encode the tokenized instruction $l$ as a concatenation of one-hot vectors. For each $i$, we categorize the tokens that appear at $l_i$ to produce a one-hot vector $h_i$ for each instruction. Each instruction $l$ is thus encoded as $L(l)=[h_1;h_2;h_3]$. This uniquely represents each instruction as a vector to condition the goal-generating policy.

\begin{table}[t!]
\resizebox{\columnwidth}{!}{

\begin{tabular}{ |p{1.34cm}|p{1.64cm}|p{1.71cm}|p{1.64cm}|p{1.82cm}|}
 \hline
 & \multicolumn{2}{|c|}{Object and landmark} & \multicolumn{2}{|c|}{Multiple objects} \\
 \hline
 Model & Error (cm) & Success (\%) & Error (cm) & Success (\%)\\
 \hline
 Pixel Flat & 7.85 $\pm$ 0.11 & 7.41 $\pm$ 0.10   & 9.31 $\pm$ 0.23   & 9.44 $\pm$ 0.15 \\
 GT Flat & 6.64 $\pm$ 0.05 &  7.31 $\pm$ 0.35 & 4.63 $\pm$ 0.70   & 31.34 $\pm$ 2.80\\
 GT Hier. & 7.32 $\pm$ 0.21 & 8.05 $\pm$  0.90 &   4.75 $\pm$ 0.25   & 19.54 $\pm$ 0.50\\
 \textbf{IFIG} & \textbf{4.74 $\pm$ 0.09} & \textbf{18.3 $\pm$ 0.70} & \textbf{3.63 $\pm$ 0.02} & \textbf{32.19 $\pm$ 0.60}\\
 \hline
\end{tabular}
}
 \caption{Error and success rates, showing mean and std dev.}
 \label{tab:comparison}
\end{table}

\subsection{Baseline approaches}
\label{baselines}

In this section we describe the baseline approaches we use to benchmark the performance of IFIG. Policies are approximated by a neural network, with reward computed according to Equation \ref{reward} for instruction-conditioned policies. \textit{Ground-truth states} are 6-d vectors containing the $xy$ positions of the end-effector and both objects. \textit{Raw-pixel states} are the raw 48$\times$48$\times$3 RGB images of the scene.

\subsubsection{Flat architecture with raw-pixel state}

To study the effectiveness of goal-setting, we compare an IFIG agent to a visual agent that does not use goal-setting. Previous approaches use flat architectures to map states directly to actions without sub-goals \cite{mei2016listen,misra2017mapping,hermann2017grounded}. We implement an architecture (shorthand: \textit{pixel flat}) that learns to map an instruction and raw visual state to an action. This is implemented by mapping the raw pixel image to a 6-d vector embedding (like our encoder), concatenating the visual embedding with the instruction embedding, and mapping the joint representation to an action through fully connected layers.

\subsubsection{Flat architecture with ground-truth state}

To further study the effectiveness of goal-setting, we compare an IFIG agent to a flat agent with access to ground-truth object positions (shorthand: \textit{GT flat}). Providing ground-truth state simplifies the instruction-following problem by eliminating the need to learn representations from high-dimensional images. This is implemented by concatenating the instruction embedding and the 6-d ground-truth state vector, and mapping it directly to an action.

\subsubsection{Hierarchical architecture with ground-truth state}

In order to study the effectiveness of the visual modality for goal-setting, we compare an IFIG agent to a similar architecture that uses ground-truth states instead of latent visual states (shorthand: \textit{GT hierarchical}). This agent learns to reach arbitrary goal states using the SAC algorithm, and learns to generate goal states from instructions and states according to the training procedure in Section \ref{goalgenpol}. Since ground-truth states describe the positions of entities in Cartesian space, extrinsic reward is computed based on the Euclidean distance between current state $s_t$ and goal state $s_g$, given as $R(s_t,s_g)=-||s_t-s_g||_2$.

\subsection{Hyperparameters}
\label{hyperparams}

We set latent space dimensionality to 6, to capture the 6 factors of variation in the positions of the end-effector and both objects. The VAE is a 3-layer convolutional neural network (CNN) with kernel sizes 5x5, 3x3, 3x3; number of output filters 16, 32, 64; and strides 3, 2, 2, respectively, followed by two fully connected layers. Controller is a 2-layer MLP with 400 and 300 hidden units, respectively. Meta-controller is a 2-layer MLP with 128 hidden units in each layer. Path length is 25 for controller and 10 for meta-controller. Batch size is 64, discount factor is 0.99, and policy learning rate, value function learning rate, and target update rate are each 0.005.

%mapping the flattened convolutional features to $\mu$ and $\sigma$

Both networks in GT hierarchical are 2-layer MLPs with 256 hidden units and 128 hidden units, respectively, and are trained via SAC with learning rate 0.005 and batch size 64. GT flat is a 2-layer MLP with 256 hidden units, trained via SAC with learning rate 0.003 and batch size 64. Pixel flat is a 3-layer CNN followed by two fully connected layers of size 64, with kernel sizes: 5x5, 3x3, 3x3; number of output filters: 32, 64, 64; and strides: 3, 1, 1, trained via SAC with learning rate 3e-4 and batch size 64. The instruction embedding is concatenated along with the flattened convolutional features, which are then sent through the fully connected layers.

%\begin{figure}[t!]
    %\centering
    %\includegraphics[width=0.45\textwidth]{fig_traj.png}
    %\caption{Sample trajectories of a blue puck, green puck, and arm %for the instruction \textit{move green puck away from blue puck}. %The sequence of latent goal reconstructions, raw pixel images of %reached states, and ground truth object positions during %execution are labeled 0-3. Goals describe the sequence of moving %to the green puck and increasing its distance to the blue puck.}
%    \label{fig:fig_traj}
%\end{figure}

\section{Results and Discussion}

\begin{figure}[t]
    \centering
    \includegraphics[width=0.47\textwidth]{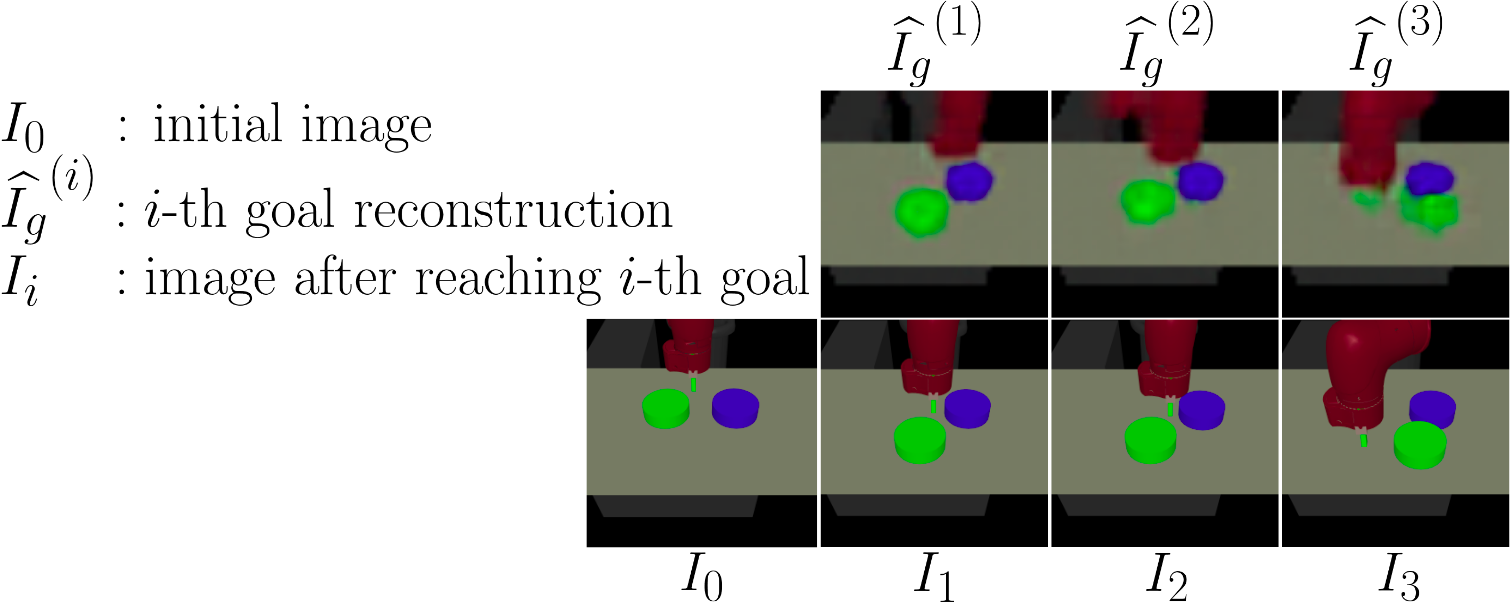}
    \caption{Example policy execution for the instruction \textit{``move the green puck North of the blue puck''}.}
    \label{fig:fig_policyexec}
\end{figure}

Our architecture outperforms all three baseline approaches in each environment. Training error and success rates for each approach are summarized in Figure \ref{fig:fig_results} and Table \ref{tab:comparison}. Success rate is measured using a threshold of 0.025 m. Selected visual goal examples are visualized in Figures \ref{fig:fig_instrs_all} and \ref{fig:fig_policyexec}.

\subsubsection{Effectiveness of goal-setting} Our IFIG agent achieves 2.47 and 3.41 times the success rate of the pixel flat agent (in the single and multiple object environments, respectively), while having 0.97 times the parameter count. Moreover, IFIG achieves, respectively, 2.50 and 1.03 times the success rate of the GT flat agent, which is given ground truth knowledge that IFIG does not require. Thus, setting visual goals appears to be cost-effective in our experiments on visual object arrangement.

\subsubsection{Latent visual structure as a sub-goal modality}

Without access to ground-truth state, our architecture achieves 2.27 and 1.65 times the success rate of the GT hierarchical agent in each environment, respectively. Controller performance appeared to be a bottleneck for both models. The IFIG controller achieves a final error of 0.9 times that of the GT hierarchical controller, consistent with the results from \cite{pong2019skew}. From this we can gather that latent visual structure provides an effective sub-goal modality for object arrangement tasks.

\subsubsection{Qualitative analysis of imagined goals}

In Figure \ref{fig:fig_policyexec}, we visualize an example execution of a trained IFIG agent, showing the sequence of visual goals generated in the successful execution of an instruction. The goal sequence shows the decomposition of an object arrangement task into a series of sub-goal arrangements. We can observe that the model has learned to localize an object by reference, imagine changes to its position in the environment according to the instruction by generating goals in latent space, and successfully reach each goal in sequence. Our results suggest a promising direction for future research on improved methods using visual goals for instruction-following.

%In Figure \ref{fig:fig_instrs_all} we visualize some of the goals produced by IFIG at the end of training, which appear qualitatively coherent, appearing to respect the rules of object permanence, constancy of the puck size, and perspective. Since the VAE is not trained with visual priors or visual coherency objectives, the induced latent distribution allows the agent to sometimes imagine goals that are slightly incoherent, as those shown in Figure \ref{fig:fig_incoherent}. While IFIG outperforms the baseline approaches despite this shortcoming, in future work, it may be useful to enhance the latent representations for improved performance and interpretability.

%\subsubsection{Policy execution} Figure \ref{fig:fig_traj} shows a policy execution of the IFIG model for the instruction \textit{move green puck away from blue puck}. In this figure, we demonstrate the agent's ability to fulfill the instruction. Before the instruction is executed, the green puck is nearby the blue puck. During policy execution, the model generates a sequence of goals that describe a path of the green puck moving increasingly farther away from the blue puck. We include a visualization of the visual goals generated by the meta-controller during this execution. The model has learned to localize an object by a reference token, and modify its position within the latent state according to the \textit{away from} spatial relation.

\section{Conclusion}

In this paper, we proposed a novel framework for training agents to execute object-arrangement instructions from raw pixel images, using hierarchical reinforcement learning to imagine and reach visual goals. The framework, called Instruction-Following with Imagined Goals (IFIG), integrates a mechanism for sequentially synthesizing visual goal states and methods for self-supervised state representation learning and visual goal reaching. IFIG requires minimal manual engineering, and assumes no prior linguistic or perceptual knowledge. The only extrinsic reward comes from a measurement of task completion. In experiments on single and multiple object arrangement, IFIG is able to decompose object-arrangement tasks into a series of visual states describing sub-goal arrangements. Without any labelling of ground-truth state, IFIG outperforms methods with access to ground-truth state in our experiments, and methods without goal-setting or explicit spatial reasoning.

\bibliographystyle{named}
\bibliography{ijcai20}

\end{document}